\pdfoutput=1
\documentclass[10pt]{article} 
\usepackage[accepted]{tmlr}

\usepackage{graphicx,amsmath,booktabs,xcolor,siunitx,microtype,nicefrac,enumitem,amsfonts,tikz}
\newcolumntype{d}[1]{S[table-format=#1]}
\definecolor{band}{RGB}{245,248,252} 

\usepackage{listings}
\usepackage[ruled,vlined,linesnumbered]{algorithm2e}

\usepackage{multirow}
\usepackage{amssymb}
\usepackage{dsfont}
\usepackage{cutwin}
\usepackage{mdframed}
\usepackage{pifont}
\usepackage[utf8]{inputenc} 
\usepackage[T1]{fontenc}
\usepackage{gensymb}
\usepackage{textcomp}
\usepackage{float} %

\usepackage[many]{tcolorbox} 
\tcbuselibrary{skins}     
\tcbuselibrary{breakable}  
\newmdenv[linewidth=0.8pt,leftmargin=0pt,topline=false,rightline=false,bottomline=false]{assbox}

\SetAlFnt{\small}
\SetAlCapFnt{\small}
\SetAlCapNameFnt{\small}
\SetAlCapHSkip{0pt}

\newcommand{\change}[1]{#1}

\usepackage[table,x11names]{xcolor} \definecolor{T2IBlue}{HTML}{E8F0FE}      
\definecolor{L2IGreen}{HTML}{E6F4EA}     
\definecolor{AgenticPurple}{HTML}{EFE8F4}  
\definecolor{OursGray}{HTML}{F5F5F5}     
\definecolor{ablationBG}{HTML}{F3F7FF} 
\definecolor{criticBG}{HTML}{FFF7F0}   
\definecolor{userBG}{HTML}{EEF8EF}     

\usepackage{color,soul}
\usepackage{caption}
\usepackage{subcaption}                        

\usepackage{wrapfig}
\usepackage{newfloat}
\DeclareCaptionStyle{ruled}{labelfont=normalfont,labelsep=colon,strut=off} 
\newcommand{\eg}{\textit{e.g.}}
\newcommand{\ie}{\textit{i.e.}}

\usepackage{hyperref}
\usepackage{cleveref}

\def\model{\textsc{CountLoop}}
\def\datasetsingle{\textsc{CountLoop-S}}
\def\datasetmulti{\textsc{CountLoop-M}}
\newcommand{\myparagraph}[1]{\vspace{0pt}\noindent{\bf #1:}}
\definecolor{DarkGreen}{RGB}{47,198,0}
\newcommand{\cmark}{\ding{51}}%
\newcommand{\xmark}{\ding{55}}%
\Crefname{equation}{Eq.}{Eqs.}
\Crefname{figure}{Fig.}{Figs.}
\Crefname{tabular}{Tab.}{Tabs.}

\usepackage{amsmath,amsfonts,bm}

\def\eqref#1{equation~\ref{#1}}

\def\1{\bm{1}}

\DeclareMathAlphabet{\mathsfit}{\encodingdefault}{\sfdefault}{m}{sl}
\SetMathAlphabet{\mathsfit}{bold}{\encodingdefault}{\sfdefault}{bx}{n}

\usepackage{url}

\title{\textsc{CountLoop}: Training-Free High-Instance Image Generation via Iterative Agent Guidance}

\author{Anindya Mondal$^{1}$, \quad Sauradip Nag$^{2,\dagger}$, \quad Ayan Banerjee$^{3,\dagger}$, \\
Josep Llad\'os$^3$, \quad Xiatian Zhu$^1$, \quad Anjan Dutta$^{1,*}$ \\
\normalfont \small
$^1$ \textit{University of Surrey} \\
$^2$ \textit{Simon Fraser University} \\
$^3$ \textit{Computer Vision Center, Universitat Aut\`onoma de Barcelona} \\
$^\dagger$ \textit{Equal Contribution} \\
$^*$ \textit{Correspondence: \texttt{ anjan.dutta@surrey.ac.uk}}
}

\def\month{08}  
\def\year{2026} 
\def\openreview{\url{https://openreview.net/forum?id=2JxXGhpCP4&invitationId=TMLR/Paper9259/}} 

\begin{document}

\maketitle

\begin{figure*}[!h]
    \centering
    \includegraphics[width=\textwidth]{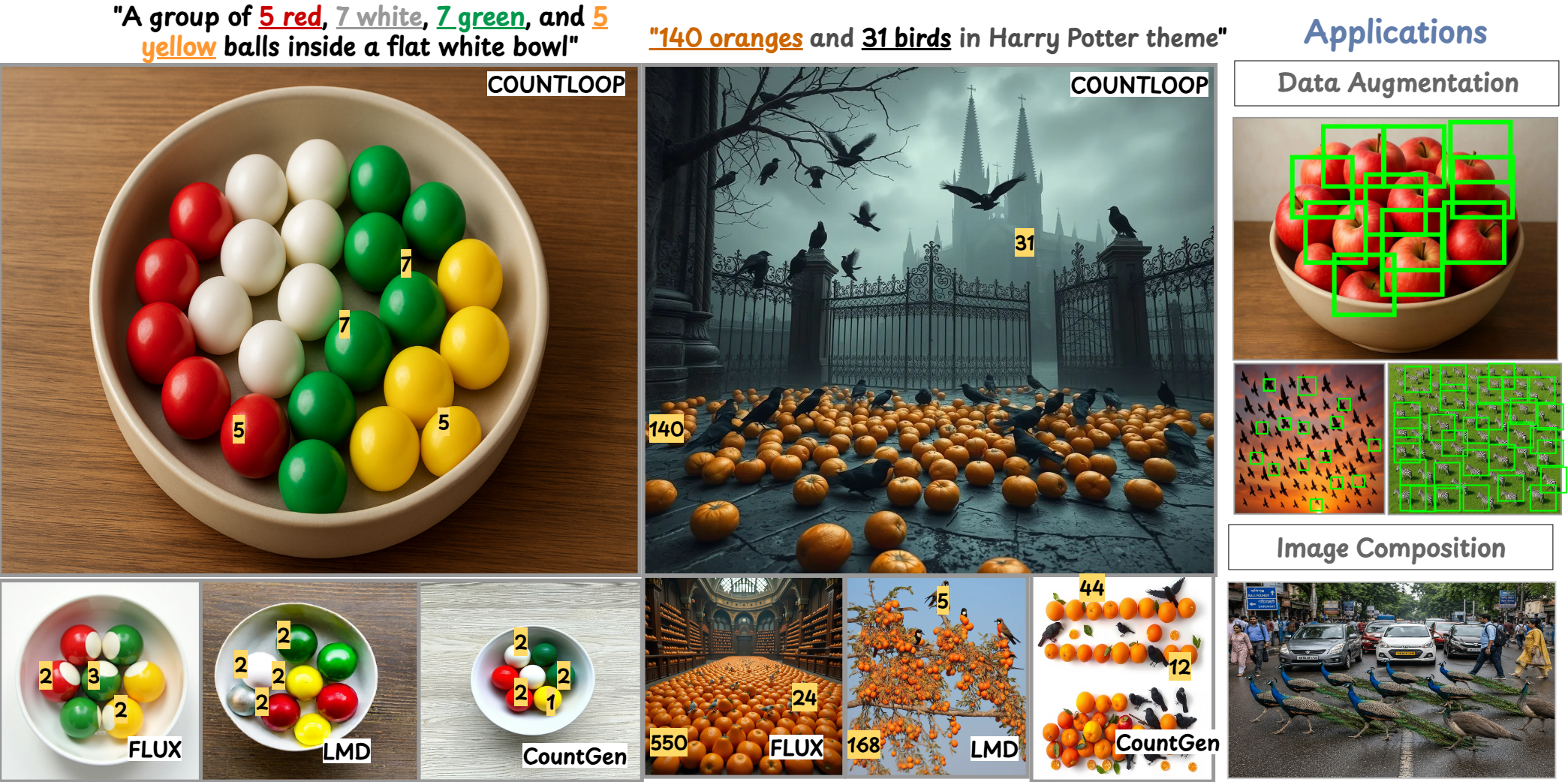}
    \caption{ Given prompts with 
    explicit per-class counts, \model~produces images whose 
    {detected counts show substantially improved alignment with targets}, even at {high} 
    cardinalities (\eg, 140 oranges and 31 birds), 
    where competing methods suffer from count saturation, semantic 
    leakage, and grid-like layouts. Unlike prior approaches, 
    \model~requires no retraining: a VLM-guided planning graph structures the layout, instance-driven attention masking prevents attribute leakage, and a Critic VLM iteratively refines the scene until the count and quality criteria are met. {Count-faithful synthesis facilitates} practical applications (right): (a) augmenting object-counting datasets~(\cite{ranjan2021learning}) with high-instance scenes, {and (b) Compositional Image Generation in complex out-of-distribution scenarios for example it is rare to find peacock in the middle of traffic
    situation.}}
    \label{fig:teaser_fin}
\end{figure*}

\begin{abstract}
Diffusion models excel at photorealistic synthesis but struggle with {object count fidelity}, especially in high-density settings. We introduce \model, a training-free framework that achieves {structured instance and count}
control through iterative, structured feedback. Our method alternates between synthesis and evaluation: a VLM-based planner generates structured scene layouts, while a VLM-based critic provides explicit feedback on
object counts, spatial arrangements, and visual quality to refine the layout iteratively. Instance-driven attention masking and cumulative attention composition further prevent semantic leakage, ensuring clear object separation even in densely occluded scenes. Evaluations on COCO-Count,
T2I-CompBench, and two newly introduced high instance benchmarks show that \model~reduces counting error by up to 57\% and achieves the highest or comparable spatial quality scores across all benchmarks, while maintaining photorealism. Project page is at \href{https://mondalanindya.github.io/CountLoop/}{https://mondalanindya.github.io/CountLoop/}.

\end{abstract}
\section{Introduction}
Digital creators, designers, and artists increasingly use text-to-image diffusion models like DALL-E 3 (\cite{betker2023improving}), SDXL (\cite{podell2023sdxl}), and FLUX (\cite{flux2024}) to produce high-quality visuals. However, these models struggle with scenes containing many distinct yet related object instances (\cite{paiss2023teaching}), limiting their effectiveness in applications where cardinality is crucial, such as {Compositional Image Generation} and augmenting object-counting datasets. Current image diffusion models typically saturate at around 10 instances per category~(\cite{binyamin2024countgen}), with {reliable cardinality control} being a known long-tail compositional failure (\cite{echo4o}), yielding semantic drift (mixed attributes), spatial collapse (cluttered or overlapping objects), or instance duplication. For instance, a prompt like ``140 oranges and 31 birds in Harry Potter theme'' might under/over-produce an incoherent pile of either oranges or birds or both ({\Cref{fig:teaser_fin}}), compromising accuracy and usability.

\begin{figure}
    \centering
    \includegraphics[width=0.8\linewidth]{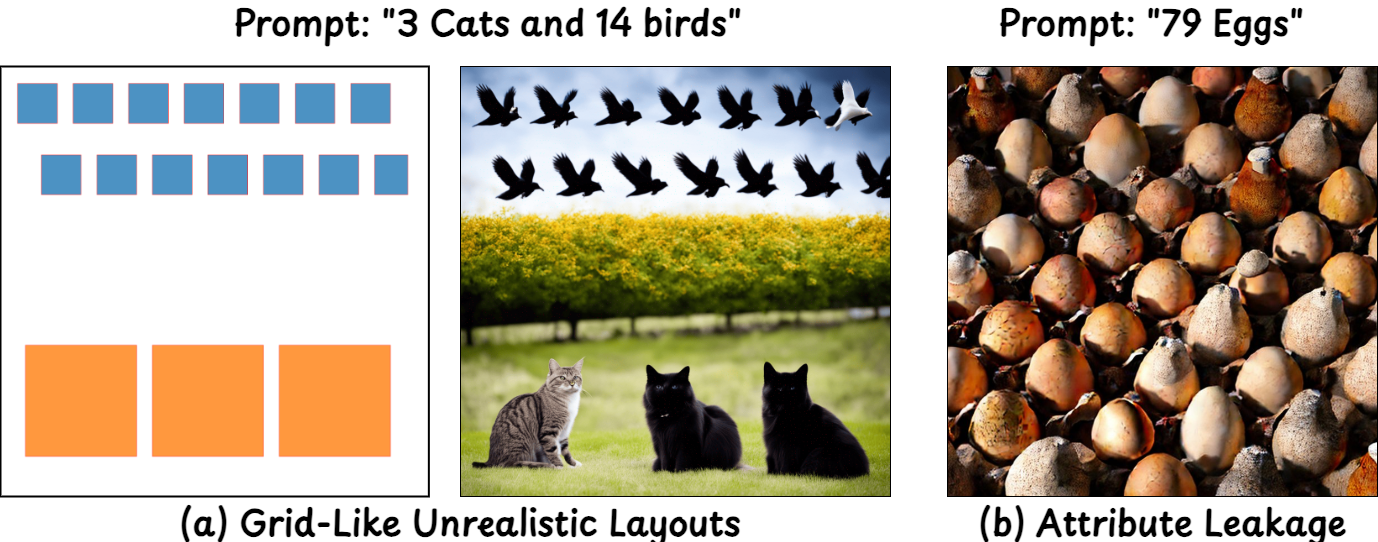}
    \caption{Issues in High-instance image generation}
    \label{fig:issues}
\end{figure}

Current solutions fall into three categories:
(1)~text-to-image (T2I) models, sometimes augmented with
gradient-based counting
guidance~(\cite{kang2023countingguidance,chefer2023attendexcite});
(2)~layout-to-image (L2I)
pipelines~(\cite{li2023gligen,feng2023layoutgpt,binyamin2024countgen,zhou2024migc,wang2024instancediffusion,zhou20253dis});
and (3)~agentic diffusion
frameworks~(\cite{wu2023selfcorrect,wang2024genartist,yang2024mastering,wu2025qwen}).
However, none scale effectively to high-instance scenes or
fully resolve the failure cases illustrated
in~\Cref{fig:issues}. Gradient-guided methods inject counting
signals during denoising but often introduce artifacts or
worsen semantic leakage as object density
increases~(\cite{dahary2024yourself,dahary2025decisive}). {Attribute or semantic leakage denotes the phenomenon where the cross-attention of the UNet lets the query tokens of one instance attend to features of other instances, causing attributes to bleed between objects. For example, in \Cref{fig:issues}(b), although the prompt asks for ``79 Eggs'', leakage distorts the cross-attention features and changes the egg textures to look like stones.} L2I pipelines guide diffusion using
bounding boxes or masks, but single-pass generation causes
cross-attention leakage, and autoregressive layout
biases~(\cite{xiong2024autoregressive}) produce unnatural,
grid-like arrangements (see \Cref{fig:issues}(a)). Agentic
frameworks use LLM-based critique but lack explicit scene
structure, leading to overcorrection or object omission, and their
focus on aesthetics over spatial precision makes them
unreliable for dense, count-sensitive generation. We present \model, a training-free framework that treats high-instance image generation as an iterative design process.
Inspired by how human designers refine compositions, \model~parses the input prompt into a planning graph encoding object attributes and spatial relationships, which guides layout-conditioned image synthesis. A VLM critic then evaluates (a)~spatial coherence and appearance fidelity via a pretrained encoder~(\cite{wu2024q}), and (b)~counting accuracy via an off-the-shelf detector — since VLMs alone struggle with precise counting in dense scenes~(\cite{visualoverload}). Critic feedback updates the planning graph and prompt, repeating until quality criteria are met.

Our \model~also introduces a cumulative attention mechanism
during the denoising process to mitigate semantic
leakage~(\cite{dahary2024yourself,dahary2025decisive}), a common
issue in high-instance scenes. Inspired by the multi-turn image generation~(\cite{cheng2024theatergen}), rather than generating
all subjects simultaneously, it synthesises one instance at a
time, providing per-instance grounding that prevents semantic
entanglement and maintains the identity of individual objects.
By imposing attention locality within instance-specific
regions~(\cite{chefer2023attendexcite,dahary2024yourself}),
\model~encourages independence across objects and prevents
the borrowing of features from nearby or similar instances.
Together, this iterative agent-guided loop, the use of
per-instance cumulative attention composition, and VLM-based
visual feedback form a training-free closed-loop pipeline. {Unlike gradient-guided methods that introduce artifacts or
L2I pipelines that produce rigid layouts, \model~acts as a
plug-and-play enhancement to standard diffusion backbones,
scaling gracefully to dense, high-instance scenes while
maintaining {substantially improved count fidelity} and natural spatial arrangements.} {While \model~may not achieve {exact counts in every dense scene}, its outputs remain highly practical for downstream tasks like image classification {and data augmentation} where image quality {and strong count alignment are paramount}. Furthermore, for {applications requiring high-precision counts}, \model~serves as an exceptionally efficient generator. \change{By drastically reducing counting error (e.g., by 43--48\% relative MAE on high-instance scenes compared to the best baselines)}, it increases the yield of {count-faithful}, compositionally coherent images that can be easily isolated via a simple post-hoc verification filter, saving substantial computational resources compared to existing baselines.}

Our contributions: \textbf{(1)}~\model, a training-free iterative pipeline for high-instance generation with {substantially improved count fidelity} and strong aesthetics; \textbf{(2)}~a cumulative attention mechanism that sequentially injects objects via instance-specific masks, mitigating semantic leakage and preserving identity in dense scenes; \textbf{(3)}~a VLM critic that evaluates count consistency and appearance fidelity, providing structured feedback to iteratively refine layout and prompt; \textbf{(4)}~evaluation on COCO-Count, T2I-CompBench, and two new high-instance benchmarks shows \model~reduces counting error by up to 57\% on standard benchmarks and 43--48\% on high-instance scenes, with the highest or comparable spatial quality across all four.

\section{Related Work}

\myparagraph{Count Control in Text-to-Image Generation}
Modern text-to-image diffusion models such as
LDM~(\cite{rombach2022ldm}), Imagen~(\cite{saharia2022imagen}),
SDXL~(\cite{podell2023sdxl}), and FLUX~(\cite{flux2024}) achieve
high photorealistic fidelity through iterative denoising, but
break down when prompts demand structured control, such as
``40 red cans on a shelf'' or ``12 apples in a bowl and 8 on
the table''. Beyond 10-15 identical objects, they often
miscount, exhibit attribute leakage, and suffer spatial
collapse~(\cite{chefer2023attendexcite,dahary2024yourself,binyamin2024countgen}).
These limitations stem from architectural constraints:
cross-attention fails to preserve per-instance identity, and
there is no global mechanism enforcing cardinality or spatial
coherence. Gradient-guided
corrections~(\cite{kang2023countingguidance,zeng2025yolo})
offer partial remedies at inference time: Counting
Guidance~(\cite{kang2023countingguidance}) steers denoising via
a regression-based counting network, while
YOLO-Count~(\cite{zeng2025yolo}) introduces a
differentiable cardinality map for token-level optimisation, improving count. However, these methods treat
counting as a global scalar constraint by optimising a signal
that tells the model {how many} objects to produce but
not {where} to place them or {how} to keep them visually distinct. As density grows, this leads to object merging and spatial collapse, fixing which requires
explicit layout structure and per-instance attention control rather than stronger counting gradients.

\myparagraph{Layout-to-Image Generation}
Layout-to-image methods condition diffusion on boxes or
masks~(\cite{li2023gligen}), LLM-derived
layouts~(\cite{lian2023llm,feng2023layoutgpt,zhao2024omost,theatron2024}), or
per-instance conditioning signals such as
instance-decomposed cross-attention with shading
aggregation~(\cite{zhou2024migc,alimohammadi2024smite}) and flexible
bbox/point/scribble inputs~(\cite{wang2024instancediffusion}).
Scene-graph pipelines~(\cite{johnson2018sg2im}) encode
pairwise relations but depend on expensive graph
annotations. More recent approaches decouple layout
planning from rendering via intermediate depth-map
synthesis~(\cite{zhou20253dis}), improving spatial coherence
across diverse backbones, while retrieval-based layout
adaptation~(\cite{binyamin2024countgen}) avoids manual
annotation but depends on retrieval coverage and the
downstream generator. However, shared limitations persist:
single-pass generation causes cross-attention leakage and
identity confusion as objects crowd
together~(\cite{chefer2023attendexcite,dahary2024yourself}),
and the absence of closed-loop feedback means counting
errors are irreversible. Robustness under high-instance
prompts ($\gg$20) remains
under-explored~(\cite{binyamin2024countgen}).

\myparagraph{Agentic Diffusion Correction}
Recent frameworks employ LLM/VLM agents as planners or critics to iteratively refine diffusion generation. SLD~(\cite{wu2023selfcorrect}) applies LLM-directed latent-space corrections (addition, deletion, repositioning) but lacks a persistent scene representation, limiting global spatial consistency. GenArtist~(\cite{wang2024genartist}) uses MLLM-driven tree planning over specialist tools, improving compositionality but not targeting high-instance count control. RPG-DiffusionMaster~(\cite{yang2024mastering}) decomposes prompts via chain-of-thought and applies regional diffusion, but rectangular non-overlapping partitions preclude dense or occluded multi-instance layouts. Qwen-Image~(\cite{wu2025qwen}) combines a VLM backbone with a diffusion transformer for strong semantic fidelity, yet lacks explicit layout conditioning and iterative correction. Despite advances in compositional control, none maintain a structured, editable scene graph, reducing reliability when {robust count control} and spatial coherence are required at high instance densities.

\section{\model}
\begin{figure*}[!t]
    \centering
    \includegraphics[width=\linewidth]{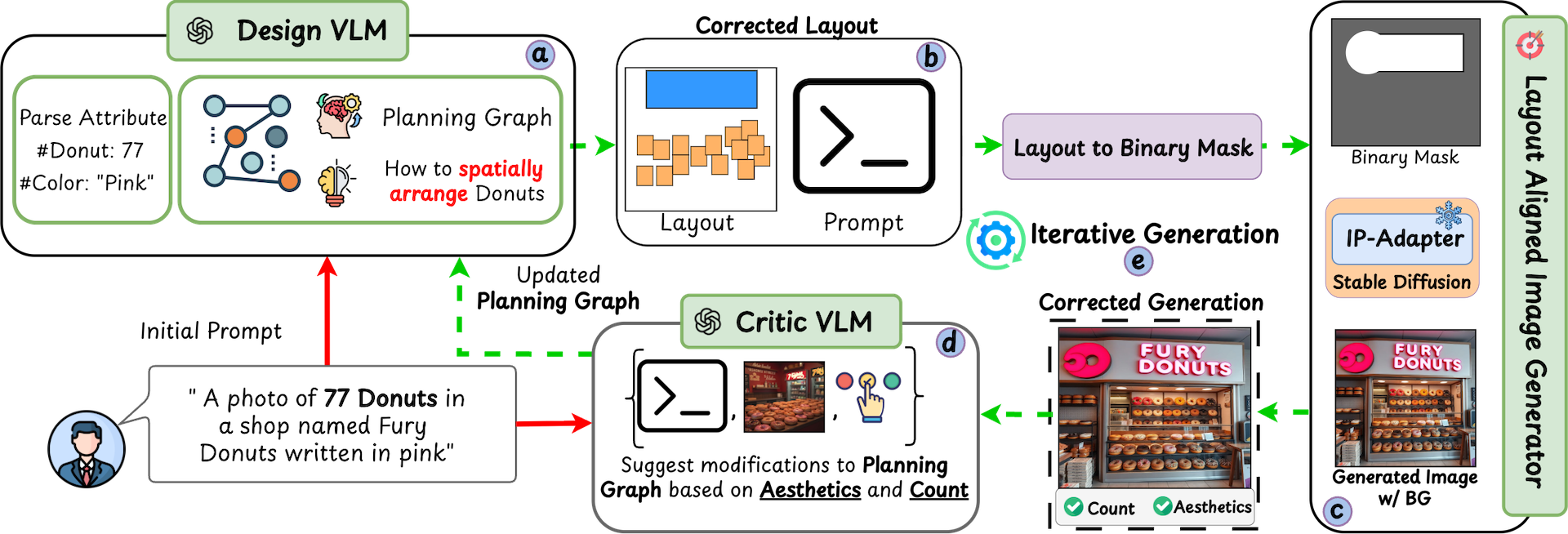}
    \caption{Given a text prompt, \textcircled{a} The Design VLM parses the prompt to construct a planning graph, which is converted into a pixel-aligned layout \textcircled{b}. \textcircled{c} This layout guides an IP-Adapter-enhanced T2I backbone for image generation. \textcircled{d} A Critic VLM evaluates the generated image's count and aesthetics, providing structured feedback to update the planning graph. \textcircled{e} This iterative loop continues until objectives are met.}
    \label{fig:pipeline}
\end{figure*}

\myparagraph{Overview}
\model~is a training-free, VLM-guided framework for high-instance image generation operating in three stages (see \Cref{fig:teaser_fin}). First, a Design VLM interprets the prompt to produce realistic, non-grid layouts (\Cref{fig:issues}(a)) with natural object placement. Second, a cumulative attention mechanism guides style-consistent generation, mitigating attribute leakage (\Cref{fig:issues}(b)) and preserving object clarity under overlap. Finally, a Critic VLM assesses counting accuracy and aesthetic quality, providing structured feedback to refine both layout and prompt.

\subsection{VLM-Guided Layout Generation}
\label{sec:layout_generation}
Precise multi-instance layout generation remains challenging: LLM-based approaches~(\cite{lian2023llm}) suffer from limited spatial reasoning~(\cite{ramachandran2025well}) and autoregressive bias, producing rigid grid-like structures (\Cref{fig:issues}(a)), while VLMs~(\cite{wu2023multimodal}) offer richer multimodal reasoning but still lack sufficient spatial flexibility. We address this by augmenting VLM Chain-of-Thought with explicit relational and spatial priors via planning graphs~(\cite{chen2024interleaved}). Building on Qwen3-VL~(\cite{yang2025qwen3}), our Design VLM produces more consistent object placement, attributes, and relations, reducing grid artifacts and yielding realistic compositions.

\myparagraph{Prompt Parsing}
As a precursor to our process, we break down the input
prompt into its core components, including object-level
quantities, instance-level attributes, and instance-level
quantities. For example, the prompt ``two cats and a bird
in the sky'' contains two objects, ``cat'' and ``bird'',
with desired quantities of two and one, respectively. The
object ``bird'' is associated with an instance-level
attribute ``in the sky'', which has a desired quantity of
one, whereas the object ``cat'' is not associated with any
instance-level attributes. We begin by instructing a VLM
(\eg, Qwen3-VL~(\cite{yang2025qwen3})) to analyze the prompt
and return a JSON dictionary. Each node carries
\texttt{id}, \texttt{category}, \texttt{pos\,[x,y]},
\texttt{size\,[w,h]}, \texttt{depth}, and \texttt{color};
edges encode \texttt{relation}, \texttt{dist}, and
\texttt{angle}; a \texttt{context} field captures the
background. These object-attribute relations serve as the
foundation for the planning graph. The full prompt schema
and a worked example are provided in the Supplementary.

\myparagraph{Planning Graph Construction} 
The graph construction process begins by using object-attribute relations parsed from the input prompt. Specifically, the planning graph is defined as $G = (V, E, B_{\text{bg}})$, where $V$ denotes object-instance nodes, $E$ represents edges encoding spatial relations, and $B_{\text{bg}}$ captures the scene context (\eg, ``outdoor environment''). Each node in $V$ includes attributes like category (\eg, cat, bird), a unique identifier (\eg, \texttt{cat\_1}), normalized position $[x, y] \in [0,1]^2$, size $[w_i, h_i]$,
depth prior $d \in [0,1]$, and color. Edges in $E$ encode spatial relations via directional operators (\eg, “above,” “left-of”), normalized distances, and angular orientations. $G$ enforces structured spatial reasoning, nodes specify individual properties while edges ensure relational consistency (\eg, minimum distances to prevent overlaps), enabling realistic multi-object scene construction. 
To integrate this structured representation into VLM reasoning, we convert the graph into a textual prompt template $P_{G}$:
\begin{equation}
\label{eq:graph}
\small{P_{G} = \phi(['Object']), ['Relation'], ['Context'])}
\end{equation}
where $\phi$ denotes a text concatenation operator; $\texttt{`Object'} \in V$, $\texttt{`Relation'} \in E$, and $\texttt{`Context'} \in B_{bg}$ denotes the textual attributes from the planning graph. Full prompt details are provided in the supplementary. The prompt $P_{G}$ encodes object positions, depth, and sizes in text, enabling spatial reasoning within the VLM, combined with in-context examples for grounding. Both $P_{G}$ and the in-context examples (denoted by $P_\text{icl}$) are fed into the Design VLM as follows:
\begin{equation}
\label{eq:design_vlm}
\mathbb{J} = \texttt{VLM}(P_{G}, P_\text{icl})
\end{equation}
where $\mathbb{J}$ is the VLM’s output in JSON format, from which we extract per-instance layouts $l_i = (x_i, y_i, w_i, h_i)$ forming the layout set $\mathbb{L} = \{l_1, \ldots, l_N\}$, the scene description prompt $P_{d}$, and background prompt $P_\text{bg}$.

\subsection{Layout Aligned Image Generation}
\label{sec:attribute_leakage}

\begin{figure}
    \centering
    \includegraphics[width=0.5\linewidth, trim=0.5cm 0.5cm 0.5cm 0.5cm, 
clip]{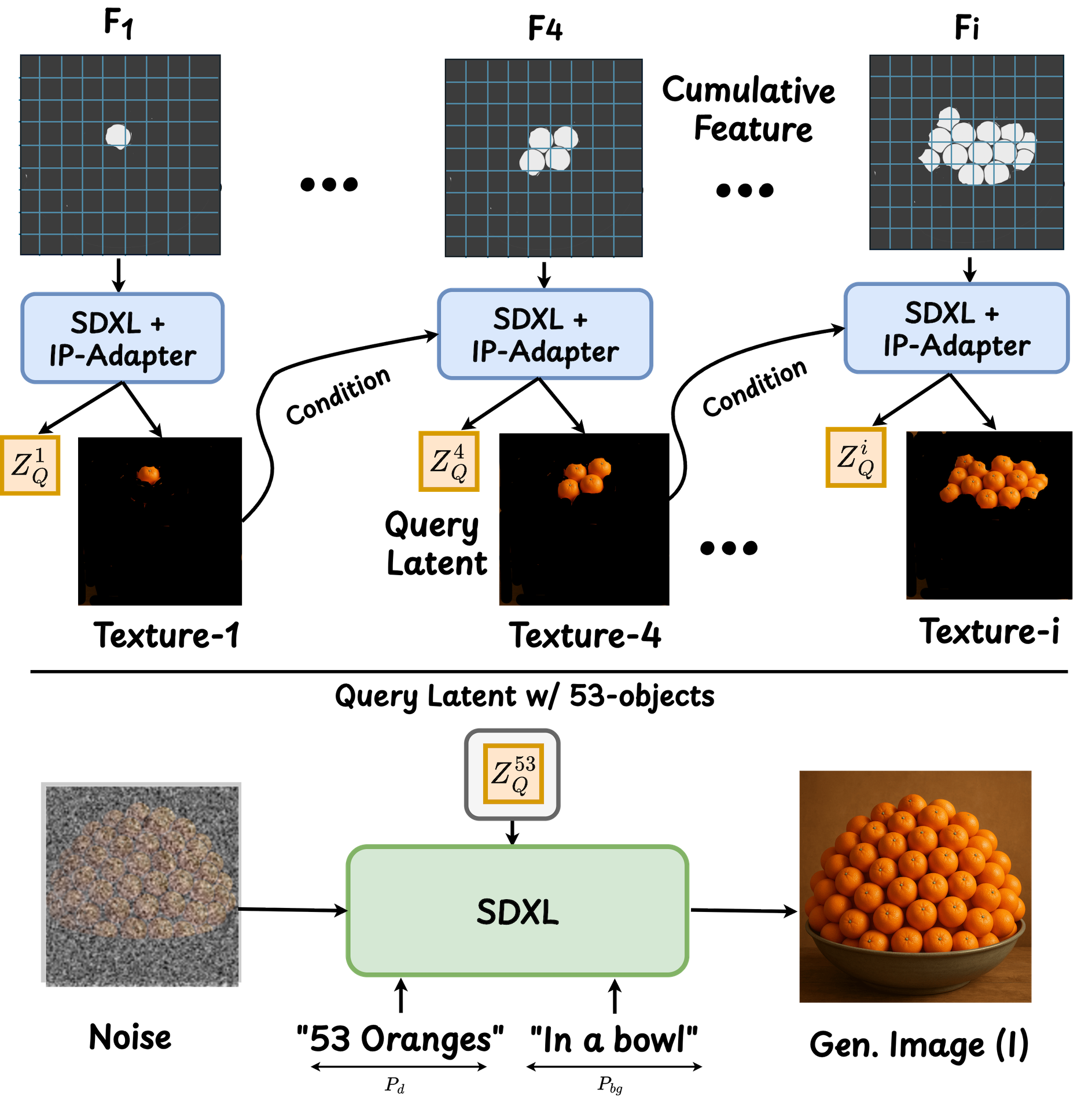}
    \caption{Cumulative latent composition, along with disentangled query feature 
extraction, mitigates attribute leakage.}
    \label{fig:cum_att}\vspace{-1.5em}
\end{figure}

Given layouts $\mathbb{L}$, layout-grounded generation commonly exhibits attribute leakage~(\cite{dahary2024yourself,dahary2025decisive}){, a phenomenon where UNet cross-attention allows query tokens of one instance to attend to features of other instances, causing attributes like color or texture to bleed between objects, }yielding correct counts but degraded visual quality (\Cref{fig:issues}(b)){; for instance, although the prompt requests ``79 Eggs'', they appear as stones due to this leakage}. Inspired by multi-turn generation~(\cite{cheng2024theatergen}), we avoid synthesising all instances in a single pass.  

\myparagraph{Layout Aligned Attention Masking}
Given the object layouts $\mathbb{L}$ and prompt description $P_d$, we aim to ground the layout with the text to generate images with {faithful} instance counts. Since layouts are discrete spatial arrangements, we project them into a continuous space using a layout encoder. 
{Following (\cite{lian2023llm}), we adapt GLIGEN(\cite{li2023gligen}) adapter (denoted by $\mathbb{E}$), 
which encodes each per-instance layout boxes $l_i \in \mathbb{L}$ into latent tokens $Q_i = \mathbb{E}(l_i)$ where $Q_{i} \in \mathbb{R}^{D}$. The full set of embeddings is represented as $Q = \{Q_1, \ldots, Q_N\}$ Since GLIGEN was built on top of SD(\cite{elegantthemes2023stable}), each layout token $Q_i$ has same dimensions $D$ as the intermediate latents of U-Net based diffusion models.}
{These tokens are
injected into the Diffusion U-Net via GLIGEN's gated self-attention mechanism in a training-free manner,
conditioning the denoising process on the layouts. During denoising, the U-Net's cross-attention layers compute spatial
attention between the latent feature map infused with the layout embeddings and the text embedding
of~$P_d$ to obtain cross-attention feature $A_{cross}$ 
thereby grounding the features with the textual description.}

However, directly using $A_{\text{cross}}$ for generation introduces semantic leakage~(\cite{dahary2024yourself}) because it attempts to generate all instances at once. To mitigate this, we independently process $A_{\text{cross}}$ at the instance level. For each object instance $i$, we apply a binary spatial mask $M_i \in \{0,1\}^{w_i \times h_i}$ (1 inside the bounding box of $l_i$, 0 elsewhere), derived from the layout $l_i \in \mathbb{L}$. {The mask is then reshaped into $\hat{M}_{i}$ using bilinear interpolation to match the latent dimension of $A_\text{cross}$}.

To obtain shape-aware instance boundaries, we refine $\hat{M}_i$ via the self-segmentation of~(\cite{dahary2024yourself}), partitioning the mask into foreground/background via $k$-means ($k{=}2$).

This produces a binary, shape-aware mask that tightly follows object contours rather than 
bounding-box boundaries. The masked layout feature is then 
computed as:
 \begin{equation}
\label{eq:mask_att}
    A^{i}_\text{mask} = A^{i}_\text{cross} \odot \hat{M}_i
\end{equation}
Here, $A^{i}_{\text{mask}}$ denotes the instance-specific masked 
attention feature, which confines the receptive field of attention 
to the corresponding object's spatial region, preventing feature 
mixing and semantic leakage across instances.

\myparagraph{Cumulative Latent Composition} 
{Once instance-level attention maps $A^i_{mask}$ have been computed for each object layout $l_i \in L$, we build the global latent feature map $F$ by sequentially placing each object’s latent features in the diffusion latent space. We initialize $F_0 = 0$ as a zero tensor in $\mathbb{R}^{H_\ell \times W_\ell \times D}$ where $H_\ell, W_\ell$ are the spatial feature dimensions and $D$ is the fixed feature dimension. Hence for $i=0,1,\ldots$ we update}

\begin{equation}
    F_{i+1}(x,y) =  \mathds{1}_{(x, y) \in l_i}\odot A_{mask}^i + (1-\mathds{1}_{(x, y) \in l_i}) \odot F_i
\end{equation}

{Here $\mathds{1}_{(x, y)}$ is the binary indicator that pixel $(x,y)$ lies within the spatial extent of $l_i$. In other words, for each pixel covered by layout $l_i$, we replace the previous feature with the new attention feature $A_{mask}^i$, and elsewhere we retain the existing feature. Because we never change the feature dimensionality in this process (each $F_i$ and $A_{mask}^i \in \mathbb{R}^D$), the dimension $D$, remains fixed (e.g. $D$ = 1280), ensuring compatibility with a frozen backbone.} {All object positioning, scale, and depth ordering are pre-specified by the layout $l_i$; we apply no further latent-space warping or scaling. In practice, instances are composed in order of increasing depth (Far $\rightarrow$ Near), so that each nearer object $i$ overwrites any existing features in its mask. 

The result is a composite latent feature map $F$ that faithfully encodes each object’s appearance, position, and scale according to the input layouts, with nearer objects occluding farther ones without any spurious blending.}

\myparagraph{Appearance Consistency via IP-Adapter} 
Generating images independently from disentangled features $F$ reduces semantic leakage but often introduces texture inconsistency, since each latent $F_i$ is denoised separately. To counter this, we condition the diffusion model (\eg, SDXL~(\cite{podell2023sdxl})) on the foreground texture of the previously generated output using IP-Adapter~(\cite{ye2023ip}). Because leakage occurs when query tokens attend to different instances during self-attention~(\cite{dahary2024yourself}), we further preserve the per-instance query representation ($Z_q$) before its interaction with keys and values, maintaining instance-level semantics. Formally:

\begin{equation}
I_{i+1}, Z^{i+1}_{q} = \Phi(F_{i+1}, P_d, \theta(I_i)), \quad i=1,\ldots,N{-}1
\end{equation}
where $I_i$ is the image generated from $F_i$, $N$ is the number of objects, and $\theta$ is IP-Adapter conditioning. The first image is generated without IP-Adapter due to the absence of prior texture. Iterating over all $F_i$ aligns prompt semantics $P_d$ with accumulated visual cues, reducing hallucinations and preserving object distinctiveness. After extracting all query embeddings $Z_q=\{Z_q^1,\ldots,Z_q^N\}$, we produce a final image with minimal attribute leakage. {To generate the final composition, we use the last query latent $Z_q^N$, which encodes all $N$ objects with consistent appearance. The attention operation is defined as: {$\mathbb{A}(Z^{N}_{q}, K, V),$}
where $K$ and $V$ are the keys and values (see ~\Cref{fig:cum_att}) of the diffusion. Each object-specific feature in $Z_q^N$ attends to a shared key–value set, enforcing semantic coherence across foreground instances while keeping the background disentangled. This operates as an implicit variant of self-attention expansion in video diffusion~(\cite{wu2023tune, alimohammadi2024smite}), but the attention is shared across object instances rather than frames. Since using only the foreground prompt $P_d$ may yield a weak background, we concatenate a dedicated background prompt $P_\text{bg}$ with $P_d$ as the textual condition to the model. The resulting image $I$ (see \Cref{fig:cum_att}) preserves the planned layout with semantically separated objects and reduced attribute leakage.}

\subsection{Layout Refinement via Iterative Feedback}
\label{sec:it_feedback}
After generating image $I$, we run an iterative refinement loop that (i)~evaluates $I$, (ii)~identifies flaws, and (iii)~updates the planning graph and prompt until count and aesthetic targets are met.

\myparagraph{Critic VLM}
We reconfigure a Qwen3-VL~(\cite{yang2025qwen3}) agent as a
Critic VLM that analyses generated images and suggests
layout revisions. Since LLM behaviour varies sharply with
instruction design~(\cite{madaan2023self,sun2023enhancing}),
the same model can serve as either creator or critic
depending on the prompt. Exploiting this, we supply a critique-style prompt $P_{\text{crit}}$ to the VLM which evaluates the generated image $I$ on two aspects: {(a) object count fidelity} and {(b) visual aesthetics}, as shown in \Cref{fig:pipeline}.
Since VLMs remain unreliable at dense
counting~(\cite{guo2025visionlanguagemodelcantcount}), {the count signal inside the Critic VLM never comes from a VLM, but rather from an off-the-shelf detection model (e.g., GroundingDINO~\cite{liu2024grounding} or CountGD++~\cite{amini2026countgd++})}, distinct from the evaluation
detector (\Cref{sec:eval}). {We note that these detector outputs serve as proxy guidance to drive refinement rather than guaranteeing absolute physical counting accuracy.} Aesthetic alignment $s_a$ is
scored by an external estimator~(\cite{wu2024q}). {A composite
score : $S = \alpha \cdot s_{c} + (1-\alpha) \cdot s_{a}$, where $s_c, s_a \in [0,1]$, is used to capture the overall quality of the generation (formulation details in the Supplementary).}
{The Critic produces structured
feedback in the form of text (denoted by $P_{\text{feed}}$) 
which is used to update the nodes and relations of the planning graph $P_{G}$, thereby 
altering the object layout size and spatial locations in the canvas. }

\begin{figure}
    \centering
    \includegraphics[width=0.8\linewidth, keepaspectratio]{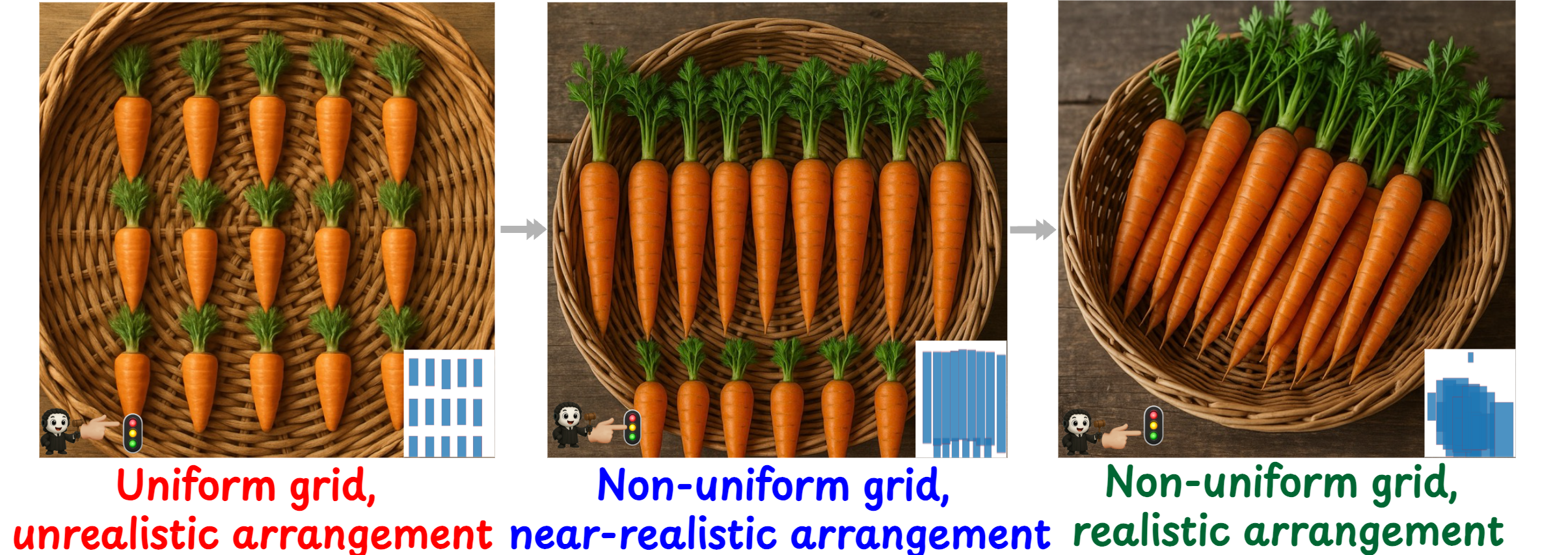}
    \caption{Successive layout refinement by Critic VLM. Layouts in the inset.}
    \label{fig:refinemet}\vspace{-1.5em}
\end{figure}

\myparagraph{Parameter-Free Refinement}
The Critic VLM's textual feedback must be translated into concrete edits to the planning graph to generate an updated image incorporating the feedback. Instead of fine-tuning
model parameters, we employ a parameter-free textual refinement
operator inspired by~(\cite{yuksekgonul2024textgrad}). We denote
this operator as $\Psi$, an LLM-based text-editing agent that
updates the planning graph through structured natural-language
reasoning. Given the current graph $G$, the critic feedback
$P_\text{feed}$, and an optimisation prompt $P_\text{opt}$,
the operator produces an updated graph:
$G' = \Psi(G, P_\text{feed}, P_{\text{opt}})$.
{Mirroring how PyTorch's AutoGrad~(\cite{paszke2017automatic})
performs backpropagation, instead of gradient tensors as feedback, the feedback here is in text form. Analogous to how backpropagation in PyTorch updates model weights, $\Psi(\cdot)$ edits (updates) the graph nodes and edges based on the textual feedback.} $\Psi(\cdot)$ interprets the input
feedback and estimates a textual analogue of a \textit{gradient},
using a loss function defined as a pre-defined textual prompt
template in $P_{\text{opt}}$. It then applies gradient-like
edits to $G$ via textual modifications rather than numerical
parameter updates.

Operating entirely on textual representations, $\Psi$ applies
targeted structural edits to $G$. For example, if the feedback identifies overlapping objects, $\Psi$ increases their spatial separation in $G$. If the feedback indicates a missing object count, $\Psi$ inserts the required missing nodes. {(A full multi-round generation case and explicit reasoning trace for this process is provided in the Supplementary.)} This parameter-free
refinement is compatible with any frozen diffusion model. After
obtaining $G'$, we derive $P_{G'}$~(~\Cref{eq:graph}) to
generate a refined layout $\mathbb{L}$~(\Cref{eq:design_vlm}),
followed by updated image synthesis $I$
(see~\Cref{fig:refinemet}). {The process terminates when the
composite score exceeds a quality threshold, \ie, {when the detector-estimated count reaches acceptable proximity to the target} and the aesthetic score is
acceptable, or after a fixed number of rounds to prevent diminishing returns from
further refinement. In practice, the majority of prompts converge
within three rounds (see convergence analysis in the
Supplementary).}

\section{Experiments}
\begin{figure*}[!t]
    \centering
    \includegraphics[width=\linewidth]{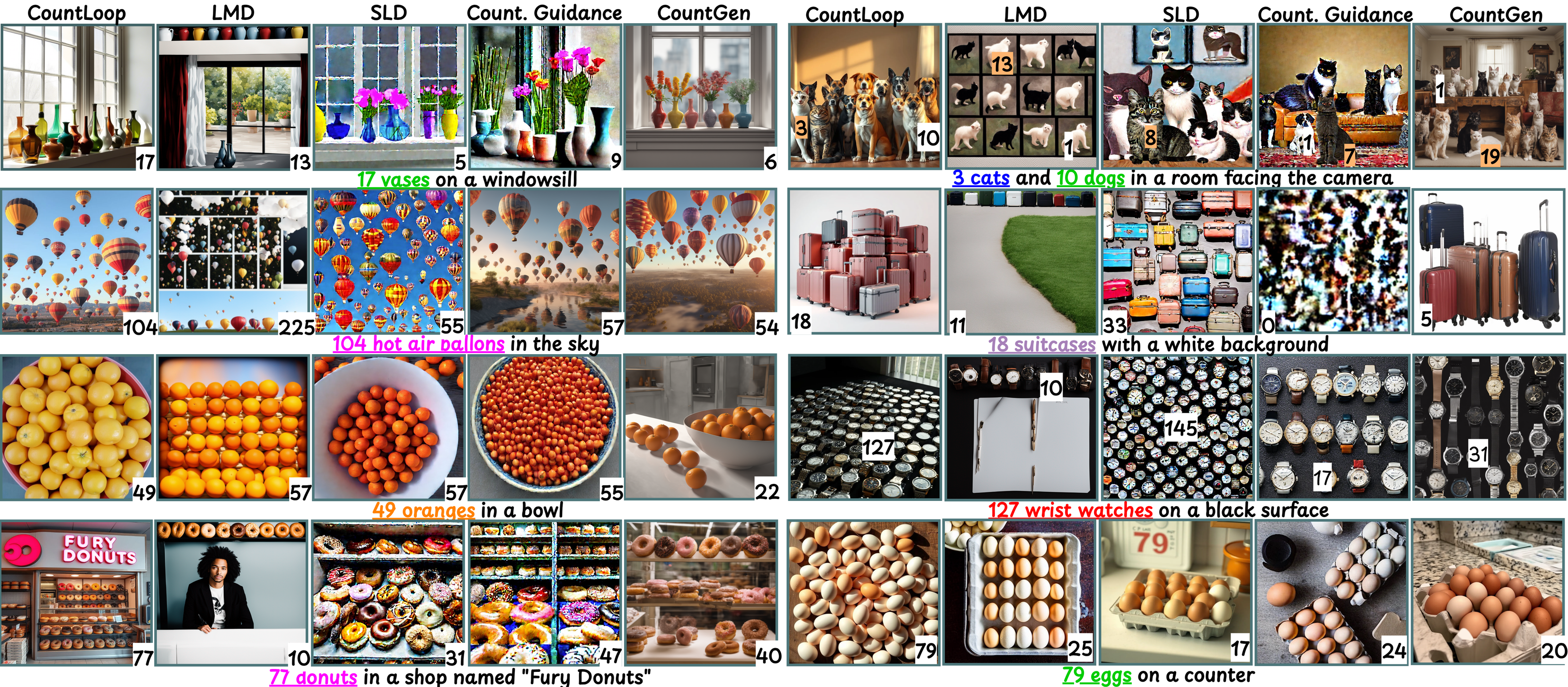}
    \caption{\model~{achieves significantly improved object count alignment} and natural arrangements in dense scenes, while methods like LMD (\cite{lian2023llm}), SLD (\cite{wu2023selfcorrect}), Counting Guidance (\cite{kang2023countingguidance}), and CountGen (\cite{binyamin2024countgen}) exhibit abnormal counts, spatial collapse, and grid artifacts. More visuals in the supplementary.}
    \label{fig:qualitative}
\end{figure*}

\subsection{Dataset and Evaluation}
\label{sec:eval}
\myparagraph{Datasets and Metric} We evaluate on four sets spanning instance count and compositional difficulty:
{COCO-Count} (MS-COCO subset (\cite{lin2014microsoft})); 
{T2I-CompBench Count} (subset of (\cite{huang2023t2i})); newly proposed {\datasetsingle} (single category, 200 prompts, 30–200 instances); and {\datasetmulti} (multi-category, 200 prompts, 30–200 instances). Benchmark construction details and examples are in the supplementary (full prompt lists released in repository). {\change{We evaluate counting performance using the MAE metric}~(\cite{kang2023countingguidance}) where OWLv2~(\cite{minderer2023scaling}) is used as an evaluator for \emph{all} methods{, configured with a confidence threshold of 0.25 and class-agnostic NMS IoU of 0.5. {Additionally, we report exact-count matches evaluated using the state-of-the-art counting model CountGD++~(\cite{amini2026countgd++}).} For counting convention, we count an object if it exceeds the minimum bounding box dimensions ($3 \times 3$) and if its center is visible (determined via the objectness token in OWLv2).} Any systematic detection bias at high densities therefore affects baselines and \model~equally, leaving relative MAE comparisons valid. {Importantly, we distinguish detector-based evaluation from true instance-count accuracy: in dense, high-cardinality scenes, automated metrics computed by off-the-shelf object detectors or density estimators (such as OWLv2 and CountGD++) serve as automated proxy evaluators rather than infallible ground truth. These vision-based evaluators can suffer from false positives, missed detections under heavy occlusion, or bounding-box overlap in crowded regions. While using identical evaluator detectors ensures fair and consistent relative comparisons across all baselines, detector-reported counts should not be conflated with guaranteed true physical counts.} {Spatial} alignment is measured via CLIP–FlanT5 encoder from VQAScore (\cite{li2024evaluating}). }


\myparagraph{Competitors} We compare \model~with representative {T2I} (SDXL~(\cite{podell2023sdxl}), FLUX~(\cite{flux2024}), SDXL-Turbo~(\cite{sauer2024adversarial}), SD3.5~(\cite{stabilityAI2025sd3.5}), Counting Guidance~(\cite{kang2023countingguidance})), {Agentic} (Qwen-Image~(\cite{wu2025qwen}), GenArtist~(\cite{wang2024genartist}), SLD~(\cite{wu2023selfcorrect}), RPG-DiffusionMaster~(\cite{yang2024mastering})), and {L2I} (LMD~(\cite{lian2023llm}), MIGC~(\cite{zhou2024migc}), CountGen~(\cite{binyamin2024countgen}), 3DIS~(\cite{zhou20253dis}), InstanceDiffusion~(\cite{wang2024instancediffusion})) methods. {Note that LMD and CountGen generate layouts with their own native modules. Box-conditioned methods without a native text-to-layout stage (MIGC, InstanceDiffusion, 3DIS) receive layouts from LMD's LLM layout generator, not from our planning graph. Thus, no baseline benefits from or is handicapped by our layout prior.} 
Implementation details are provided in the supplementary.
\subsection{Main Results}

\myparagraph{Quantitative Results}
\Cref{tab:count_clip_split} reports counting error (MAE, lower
is better), {exact-count match (Ex.$\uparrow$)}, and spatial quality across all benchmarks. In the
low-instance regime of COCO-Count, \model~achieves the lowest
overall MAE (0.45), outperforming strong agentic competitors
such as Qwen-Image (1.04) and SLD (1.15). {These gains are modest in absolute terms because existing methods already perform well at low counts, where the failure modes \model~targets, like semantic leakage, layout
rigidity, and count saturation, have not yet manifested}. The critical
differentiator emerges at scale: on \datasetsingle,
\model~records a MAE of 7.59, less than half the error of the strongest agentic competitor (Qwen-Image: 17.30) and the best L2I baseline (3DIS: 14.55). Methods that perform competitively at low counts suffer clear collapse at scale: CountGen's MAE rises from 1.88 to 34.44 and SLD's from 1.15 to 29.65. Even recent baselines evaluated only on \datasetsingle~(InstanceDiffusion: 16.07, Qwen-Image: 17.30) remain $2{\times}$ above \model. This robustness extends to multi-category scenes (\datasetmulti, MAE: 2.13). {This capability is further highlighted by the exact-count match (Ex.): \change{\model~achieves 20\% on \datasetsingle~and 33\% on \datasetmulti~(rising to 29\% and 42\% with the CountGD++ critic), outperforming the best baseline (3DIS at 6\% and 27\%, respectively)}. {However, as exact-match rates remain low across all models in high-density regimes ($N \ge 30$), the evidence does not establish guaranteed or exact count control at extreme cardinalities, where occlusions and canvas resolution limits inherently constrain exact matching; rather, \model~substantially narrows counting error and improves tolerance alignment.}} Notably, \model~achieves this without sacrificing generation quality: its spatial score on \datasetsingle~is 0.93 versus 0.75 (SLD) and 0.74 (Qwen-Image), demonstrating that the Critic VLM resolves the count-quality trade-off that constrains all prior paradigms. {To provide a commercial reference, we evaluated a 100-prompt CountLoop-S subsample on Nano Banana Pro, which yielded an MAE of 19.06 and spatial score of 0.70 (compared to our 7.59/0.93), demonstrating that even commercial systems saturate in the 15--20 instance range.}

\begin{table*}[!t]
\centering \scriptsize
\caption{\textbf{Counting and spatial quality across benchmarks.}
We report counting error (MAE$\downarrow$), {exact-count match (Ex.$\uparrow$)}, and spatial quality
(Spatial$\uparrow$) for single-category and multi-category prompts.
{Agentic baselines are run with their native self-correction loops fully enabled.}
\textbf{Best} in bold, \underline{second-best} underlined.}
\label{tab:count_clip_split}
\setlength{\tabcolsep}{4pt}
\renewcommand{\arraystretch}{1.15}
\setlength{\aboverulesep}{0pt}
\setlength{\belowrulesep}{0pt}
\resizebox{\textwidth}{!}{%
\begin{tabular}{@{}ll ccc ccc ccc ccc@{}}
  \toprule
  \multirow{2}{*}{\textbf{Family}} & \multirow{2}{*}{\textbf{Model}} &
  \multicolumn{6}{c}{\textbf{Single Category}} &
  \multicolumn{6}{c}{\textbf{Multi Category}} \\
  \cmidrule(lr){3-8}\cmidrule(l){9-14}
  & &
  \multicolumn{3}{c}{\textbf{COCO-Count}} &
  \multicolumn{3}{c}{\textbf{T2I-CompBench}} &
  \multicolumn{3}{c}{\textbf{\datasetsingle}} &
  \multicolumn{3}{c}{\textbf{\datasetmulti}} \\
  \cmidrule(lr){3-5}\cmidrule(lr){6-8}\cmidrule(lr){9-11}\cmidrule(l){12-14}
  & & MAE$\downarrow$ & {Ex.$\uparrow$} & Sp.$\uparrow$
    & MAE$\downarrow$ & {Ex.$\uparrow$} & Sp.$\uparrow$
    & MAE$\downarrow$ & {Ex.$\uparrow$} & Sp.$\uparrow$
    & MAE$\downarrow$ & {Ex.$\uparrow$} & Sp.$\uparrow$ \\
  \midrule
  \rowcolor{T2IBlue}
  & SDXL~(\cite{podell2023sdxl})
    & 2.37 & {25\%} & 0.38 & 2.72 & {25\%} & 0.75 & 29.96 & {0\%} & 0.63 & 9.89 & {10\%} & 0.55 \\
  \rowcolor{T2IBlue}
  & FLUX~(\cite{flux2024})
    & 1.40 & {55\%} & 0.53 & 1.48 & {55\%} & \underline{0.78} & 17.47 & {4\%} & 0.65 & 9.62 & {12\%} & 0.58 \\
  \rowcolor{T2IBlue}
  & SD 3.5~(\cite{stabilityAI2025sd3.5})
    & 1.10 & {55\%} & 0.46 & 1.58 & {40\%} & 0.76 & 21.81 & {0\%} & 0.64 & 8.40 & {16\%} & 0.56 \\
  \rowcolor{T2IBlue}
  & SDXL-Turbo~(\cite{sauer2024adversarial})
    & 2.50 & {25\%} & 0.23 & 3.76 & {10\%} & 0.53 & 51.14 & {0\%} & 0.39 & 9.95 & {10\%} & 0.37 \\
  \rowcolor{T2IBlue}
  \multirow{-5}{*}{\textbf{T2I}} & Counting Guidance~(\cite{kang2023countingguidance})
    & 1.68 & {40\%} & 0.63 & 3.90 & {10\%} & 0.56 & 42.49 & {0\%} & 0.47 & 8.43 & {16\%} & 0.41 \\
  \midrule
  \rowcolor{L2IGreen}
  & LMD~(\cite{lian2023llm})
    & 3.09 & {10\%} & 0.24 & 5.56 & {2\%} & 0.73 & 16.62 & {5\%} & 0.66 & 6.34 & {25\%} & 0.64 \\
  \rowcolor{L2IGreen}
  & MIGC~(\cite{zhou2024migc})
    & 1.83 & {40\%} & 0.36 & 2.96 & {25\%} & 0.65 & 17.54 & {4\%} & 0.65 & 6.28 & {25\%} & 0.62 \\
  \rowcolor{L2IGreen}
  & CountGen~(\cite{binyamin2024countgen})
    & 1.88 & {40\%} & 0.61 & 5.22 & {2\%} & 0.75 & 34.44 & {0\%} & 0.72 & 6.46 & {24\%} & \underline{0.69} \\
  \rowcolor{L2IGreen}
  & InstanceDiffusion~(\cite{wang2024instancediffusion})
    & 1.77 & {40\%} & 0.40 & 2.83 & {25\%} & 0.68 & 16.07 & {5\%} & 0.74 & 6.11 & {26\%} & 0.66 \\
  \rowcolor{L2IGreen}
  \multirow{-5}{*}{\textbf{L2I}} & 3DIS~(\cite{zhou20253dis})
    & 1.56 & {40\%} & 0.42 & 2.56 & {25\%} & 0.70 & \underline{14.55} & {6\%} & \underline{0.76} & 5.75 & {27\%} & \underline{0.69} \\
  \midrule
  \rowcolor{AgenticPurple}
  & GenArtist~(\cite{wang2024genartist})
    & 1.50 & {40\%} & 0.45 & 1.50 & {40\%} & 0.70 & 32.47 & {0\%} & 0.60 & 4.93 & {11\%} & 0.57 \\
  \rowcolor{AgenticPurple}
  & SLD~(\cite{wu2023selfcorrect})
    & 1.15 & {55\%} & \underline{0.70} & 1.44 & {55\%} & 0.77 & 29.65 & {0\%} & 0.75 & \underline{3.74} & {20\%} & 0.65 \\
  \rowcolor{AgenticPurple}
  & RPG~(\cite{yang2024mastering})
    & 1.28 & {55\%} & 0.60 & 1.47 & {55\%} & 0.75 & 31.85 & {0\%} & 0.70 & 4.34 & {15\%} & 0.62 \\
  \rowcolor{AgenticPurple}
  \multirow{-4}{*}{\textbf{Agentic}} & Qwen-Image~(\cite{wu2025qwen})
    & \underline{1.04} & {55\%} & 0.58 & \underline{1.26} & {55\%} & 0.77 & 17.30 & {4\%} & 0.74 & 6.92 & {22\%} & 0.66 \\
  \specialrule{0.8pt}{0pt}{0pt}
  \rowcolor{OursGray}
  \multirow{-1}{*}{\textbf{Ours}} & \textbf{\model~(w/ GDINO critic)}
    & 0.45 & {85\%} & 0.93
    & 1.23 & {55\%} & 0.79
    & 7.59 & {20\%} & 0.93
    & 2.13 & {33\%} & 0.73 \\
  \rowcolor{OursGray}
  & {\textbf{\model~(w/ CountGD++ critic)}}
    & {\textbf{0.32}} & {\textbf{85\%}} & {\textbf{0.98}}
    & {\textbf{0.87}} & {\textbf{70\%}} & {\textbf{0.95}}
    & {\textbf{5.33}} & {\textbf{29\%}} & {\textbf{0.97}}
    & {\textbf{1.02}} & {\textbf{42\%}} & {\textbf{0.85}} \\
  \bottomrule
\end{tabular}}
\end{table*}

\myparagraph{Qualitative Results}  
\Cref{fig:qualitative} demonstrates \model's {superior counting fidelity} across diverse instance counts. For ``17 vases'', competitors under-generate (LMD: 13, Count Guidance: 9, CountGen: 6), while \model~accurately renders all 17 with natural arrangements. In the ``104 hot air balloons'' scene, \model~{faithfully} places balloons with realistic spacing {and much closer count fidelity}, unlike Count Guidance (57), CountGen (54), and LMD's artificial clusters (225 overlapping). 
\model~consistently avoids semantic drift, grid artifacts, and count inaccuracies that plague baselines, outperforming all competitors on high-instance scenes.

\subsection{Ablations and Analysis}
\begin{figure}[h]
    \centering
    \begin{subfigure}[t]{0.262\linewidth}
        \centering
        \includegraphics[width=\linewidth]{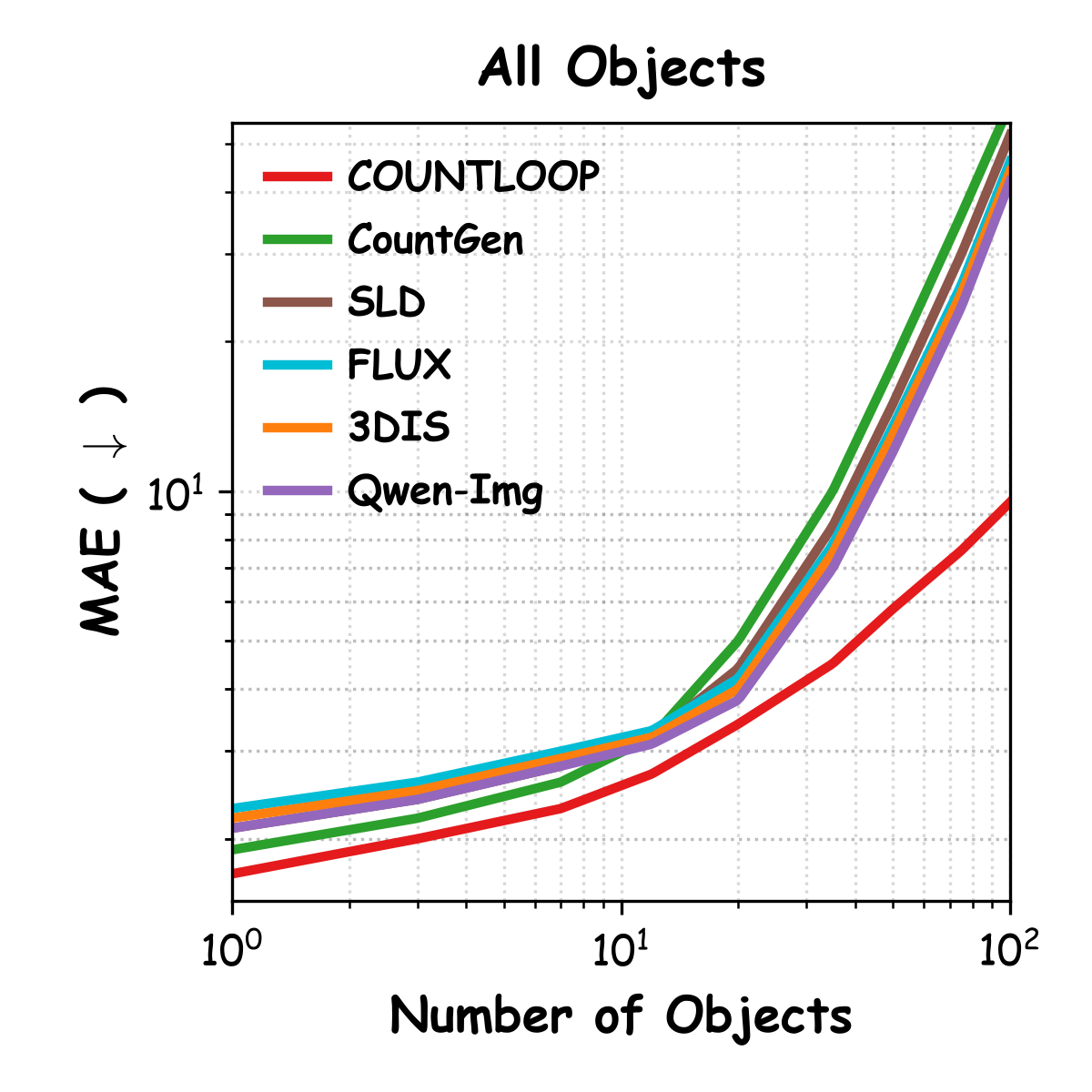}
        \caption{\change{MAE vs.\ object count.}}
        \label{fig:mae_count}
    \end{subfigure}%
    \hfill
    \begin{subfigure}[t]{0.722\linewidth}
        \centering
        \includegraphics[width=\linewidth]{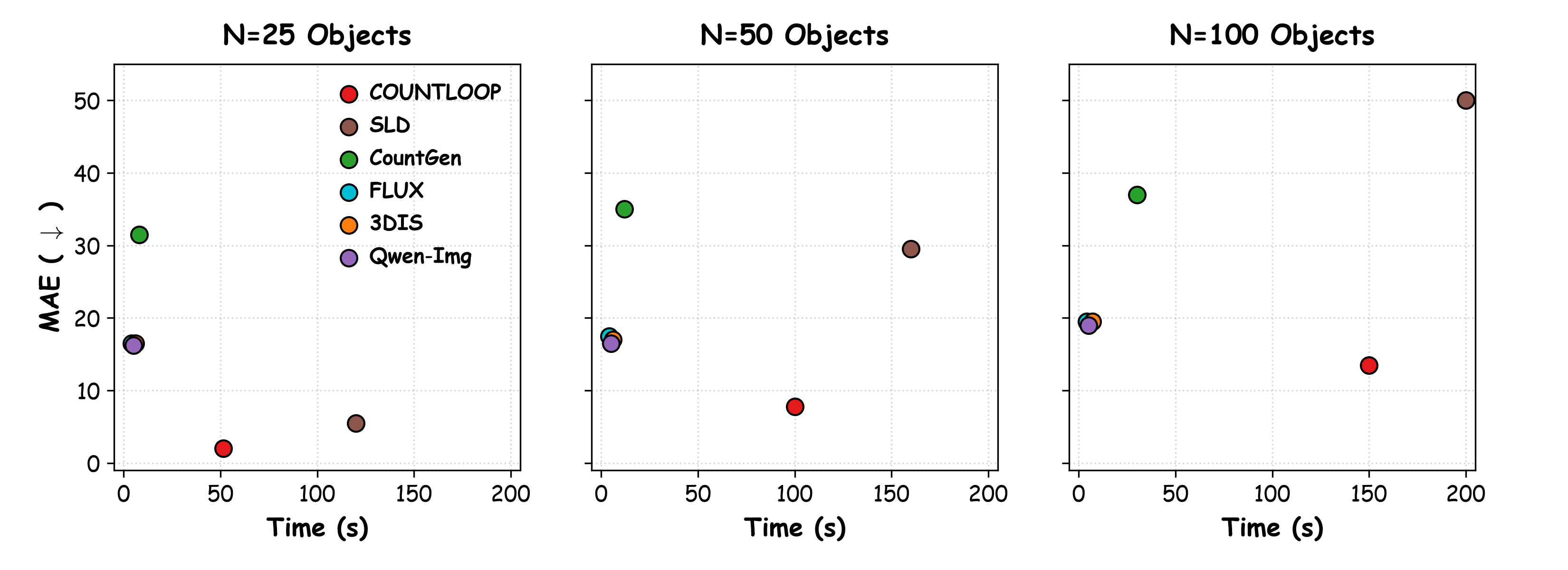}
        \caption{\change{MAE vs.\ runtime.}}
        \label{fig:anytime_runtime}
    \end{subfigure}
    \caption{\change{\textit{Left:} Counting error (MAE) rises with instance count.
             \textit{Right:} Runtime curves echo the same ordering.}}
    \label{fig:count_runtime_alignment}
\end{figure}

\myparagraph{Key Components}
\Cref{tab:ablation_study_components} progressively builds \model~from a plain baseline to quantify the contribution of each component on \datasetsingle. Both the Planning Graph (\textbf{PG}) and Cumulative Attention (\textbf{CA}) independently halve the counting error relative to the baseline (MAE: 29.91$\rightarrow$ 14.98 and 14.39, respectively), confirming that structured prompt decomposition and leakage-free attention each address a distinct and roughly equal source of failure. {As shown, layout grounding \change{reduces counting error (MAE) by 50\%} compared to mere text conditioning alone.} Combining them (PG+CA: 11.27) yields further gains{, outperforming standard single-pass L2I methods (e.g., 3DIS at 14.55) by structuring the prompt before generation.} Iterative Refinement (\textbf{IR}) closes the remaining gap with a 33\% additional MAE reduction (11.27 $\rightarrow$ 7.59), recovering instances that a single forward pass cannot place correctly. {A random-retry baseline (best of 3 attempts with new seeds) only reaches 10.90, confirming that directed feedback—not merely repetition—drives these gains. Notably, pairing our IR critic with a standard generation backbone (simulated by dropping CA) causes errors to double (7.59 $\rightarrow$ 14.98), proving the critic alone is insufficient without instance-driven attention.}
{We also show that beyond a critical instance count, the physical area per instance on a fixed-resolution canvas becomes too small for objects to remain visually distinguishable, which is a limit shared by all generation methods and that \model~reaches this floor at substantially higher $N$ than
competitors (\Cref{fig:mae_count})}.

{\myparagraph{Runtime Analysis}
We evaluate end-to-end runtime on \datasetsingle~by measuring MAE vs.\ wall-clock time at $N \in \{25, 50, 100\}$ (\Cref{fig:anytime_runtime}). One-shot baselines (FLUX~(\cite{flux2024}), Qwen-Image~(\cite{wu2025qwen})) are fast ($<10$\,s) but incur high, irreducible error. L2I methods invest moderate compute yet plateau at comparable error. Among iterative methods, \model~dominates SLD~(\cite{wu2023selfcorrect}) at every scale: at $N{=}100$, $3{\times}$ lower error (MAE~$\approx 13$ vs.\ $50$) in $25\%$ less time (${\sim}150$\,s vs.\ $200$\,s); \change{at $N{=}25$, \model~achieves MAE~$\approx 2$ in ${\sim}50$\,s while SLD requires ${\sim}120$\,s to reach MAE~$\approx 5.5$ (whereas one-shot baselines plateau around MAE~$\approx 17$)}. This mirrors \Cref{fig:mae_count}, where competing methods plateau beyond ${\sim}10$--$20$ objects. {We detail the {comprehensive} per-stage runtime breakdown in the supplementary material.}} 

\begin{table}[t]
\centering
\setlength{\tabcolsep}{4pt}
\renewcommand{\arraystretch}{1.12}
\setlength{\aboverulesep}{0pt}
\setlength{\belowrulesep}{0pt}
\captionsetup[subtable]{position=top,skip=4pt}

\caption{Analysis of \model~components, critic choices, and human
evaluation. \textbf{(a)} Design-critic combinations on
\datasetsingle; the default is in \textbf{bold} and the best per
designer is \underline{underlined}. \textbf{(b, c)} Ablations of
the design components (\textbf{PG}: Planning Graph, \textbf{CA}:
Cumulative Attn., \textbf{IR}: Iterative Refinement) and critic
modules (\textbf{OVD}: Open-vocab Detector, \textbf{AS}: Aesthetic
Scorer). \textbf{(d)} User scores on a 0--5 scale (higher is better).}
\vspace{4pt}

\begin{subtable}[t]{0.56\linewidth}
\centering\small
\caption{Design--critic configurations}
\label{tab:designer_critic_grid}
\begin{tabular}{@{}l l c c@{}}
  \toprule
  \textbf{Design} & \textbf{Critic} & \textbf{MAE}$\downarrow$ & \textbf{Spatial}$\uparrow$ \\
  \midrule
  \rowcolor{T2IBlue}
  Qwen3-VL~(\cite{yang2025qwen3}) & Qwen3-VL   & \textbf{7.59}      & \textbf{0.93} \\
  \rowcolor{T2IBlue}
                                & Llava-1.5B & 12.40              & 0.70 \\
  \rowcolor{T2IBlue}
                                & Pixtral    & 11.83              & 0.72 \\
  \midrule
  \rowcolor{L2IGreen}
  Llava-1.5B~(\cite{lin2024evaluating}) & Qwen3-VL   & 11.27            & 0.74 \\
  \rowcolor{L2IGreen}
                                & Llava-1.5B & 10.85              & 0.73 \\
  \rowcolor{L2IGreen}
                                & Pixtral    & \underline{10.51}  & \underline{0.75} \\
  \midrule
  \rowcolor{AgenticPurple}
  Pixtral~(\cite{agrawal2024pixtral})  & Qwen3-VL   & \underline{10.18} & \underline{0.76} \\
  \rowcolor{AgenticPurple}
                                & Llava-1.5B & 11.05              & 0.72 \\
  \rowcolor{AgenticPurple}
                                & Pixtral    & 10.68              & 0.73 \\
  \bottomrule
\end{tabular}
\end{subtable}\hfill
\begin{subtable}[t]{0.42\linewidth}
\centering
\caption{Component ablation. {$^*$Random-retry.}}
\label{tab:ablation_study_components}
{\rowcolors{2}{ablationBG}{white}%
\begin{tabular}{@{}c c c c c@{}}
\toprule
\textbf{PG} & \textbf{CA} & \textbf{IR} & \textbf{MAE}$\downarrow$ & \textbf{Spatial}$\uparrow$ \\
\midrule
\xmark & \xmark & \xmark & 29.91 & 0.61 \\
\cmark & \xmark & \xmark & 14.98 & 0.68 \\
\xmark & \cmark & \xmark & 14.39 & 0.71 \\
\cmark & \cmark & \xmark & 11.27 & 0.81 \\
\cmark & \xmark & \cmark & 14.98 & 0.68 \\
{\cmark} & {\cmark} & {\xmark$^*$} & {10.90} & {0.81} \\
\specialrule{0.6pt}{0pt}{0pt}
\rowcolor{OursGray}
\textbf{\cmark} & \textbf{\cmark} & \textbf{\cmark} & \textbf{7.59} & \textbf{0.93} \\
\bottomrule
\end{tabular}}
\end{subtable}

\vspace{2pt}

\begin{subtable}[t]{0.38\linewidth}
\centering\small
\caption{Critic VLM configs}
\label{tab:critic_checklist_metrics}
{\rowcolors{2}{criticBG}{white}%
\begin{tabular}{@{}c c c c@{}}
\toprule
\textbf{OVD} & \textbf{AS} & \textbf{MAE}$\downarrow$ & \textbf{Spatial}$\uparrow$ \\
\midrule
\xmark & \xmark & 29.59 & 0.67 \\
\xmark & \cmark & 15.49 & 0.70 \\
\cmark & \xmark &  9.27 & 0.83 \\
\specialrule{0.6pt}{0pt}{0pt}
\rowcolor{OursGray}
\cmark & \cmark & \textbf{7.59} & \textbf{0.93} \\
\bottomrule
\end{tabular}}
\end{subtable}\hfill
\begin{subtable}[t]{0.60\linewidth}
\centering\small
\caption{User evaluation}
\label{tab:user_study}
{\rowcolors{2}{userBG}{white}%
\begin{tabular}{@{}l c c c c c@{}}
\toprule
\textbf{Metric} & \cellcolor{OursGray}\textbf{\model} & \textbf{LMD} & \textbf{FLUX} & \textbf{SLD} & \textbf{CountGen} \\
\midrule
Alignment  & \cellcolor{OursGray}\textbf{4.5} & 3.4 & 3.7 & 4.0 & 3.6 \\
Aesthetics & \cellcolor{OursGray}\textbf{4.4} & 3.3 & 3.5 & 3.9 & 3.8 \\
Count      & \cellcolor{OursGray}\textbf{4.6} & 3.7 & 4.0 & 4.2 & 3.4 \\
Overall    & \cellcolor{OursGray}\textbf{4.5} & 3.5 & 3.7 & 4.0 & 3.6 \\
\bottomrule

\end{tabular}}

\end{subtable}
\end{table}


\myparagraph{Performance with different Design-Critic variants}
{We evaluate the impact of various Design–Critic configurations on \datasetsingle, pairing three open-source Design VLMs with three Critic VLMs under matched evaluation conditions. Results are in \Cref{tab:designer_critic_grid}. The default Qwen3-VL (as design and critic role) achieves the best performance. However, even the weakest configuration (Qwen3-VL+Llava-1.5B) still outperforms the  best competitor 3DIS(\cite{zhou20253dis}) which is training based, demonstrating that \model~is robust to VLM choice while successful in mutually guiding each other to generate plausible images.}

\myparagraph{Critic Composition}
~\Cref{tab:critic_checklist_metrics} isolates each Critic 
component. A VLM-only critic yields the weakest performance 
(MAE~29.59), confirming VLMs struggle with dense 
counting~(\cite{guo2025visionlanguagemodelcantcount}). AS alone 
substantially reduces MAE to 15.49 ($-$14.1), while OVD alone 
achieves a larger gain (MAE~9.27, $-$20.3), identifying it as 
\change{the primary driver of count fidelity}. Together they reach 
MAE~7.59, {showcasing the importance of OVD and AS in guiding our critic-VLM.}

\myparagraph{Human Evaluation} We ran a 30-participant study (20 designers, 10 AI artists) across all four benchmarks. Each participant rated 15 blinded sets of 5 images (\model, FLUX (\cite{flux2024}), LMD~(\cite{lian2023llm}), SLD (\cite{wu2023selfcorrect}), and CountGen (\cite{binyamin2024countgen})) on a 5-point scale for {Prompt Alignment}, {Aesthetic Quality}, {Count Accuracy}, and {Overall Preference}. \model~was preferred across all axes (\Cref{tab:user_study}), with clear margins over its competitors. \change{{For our human count rating, we map each rating point (0--5) to a 17\% count precision increment (e.g., 0 indicates no count alignment, 5 indicates exact match).} Human counting evaluation scores are}
consistent with OWLv2-reported trends across all methods,
providing the strongest detector-agnostic validation available:
if OWLv2 introduced systematic bias at high counts, human
rankings would diverge from detector rankings; they do not. Procedure, demographics, and the survey interface are
detailed in the supplementary.
\begin{wraptable}{r}{0.3\textwidth}
  \centering \scriptsize
  \caption{Backbone swap.}
  \label{tab:backbone_ablation}
  \setlength{\tabcolsep}{3pt}
  \resizebox{0.9\linewidth}{!}{%
    \begin{tabular}{@{}lcc@{}}
      \toprule
      \textbf{Backbone}
        & \textbf{MAE}$\downarrow$
        & \textbf{Spatial}$\uparrow$ \\
      \midrule
      \rowcolor{AgenticPurple}
      SD v1.5 & 8.05 & 0.88 \\
      \rowcolor{AgenticPurple}
      SD 3.5  & 7.44 & 0.90 \\
      \rowcolor{AgenticPurple}
      SDXL    & \textbf{7.59} & \textbf{0.93} \\
      \bottomrule
    \end{tabular}%
  }
\end{wraptable}

\myparagraph{Performance across different T2I backbones} To assess the generality of \model~across diffusion backbones, we replaced the default SDXL model with two additional Stable Diffusion checkpoints: \textit{SD v1.5} and \textit{SD 3.5}. We kept all other components (planning graph, cumulative attention, IP-Adapter, critic loop) and hyperparameters identical. \Cref{tab:backbone_ablation} reports counting MAE, and spatial scores on the \datasetsingle~benchmark. While all backbones benefit substantially from \model's structured refinement, we observe that higher-capacity models yield marginally better spatial coherence, with SDXL at the top. Counting performance remains stable (MAE $\leq$ 8.1) across all three backbones, indicating that \model's instance-control mechanism is largely model-agnostic.

\myparagraph{Robustness to Critic Components}
The Critic VLM relies on two external modules: an
open-vocabulary detector (OVD) for count verification and
an aesthetic scorer (AS) for visual quality assessment. By
default, we use the base
GroundingDINO~(\cite{liu2024grounding}) checkpoint as the OVD
and Q-Align~(\cite{wu2024q}) as the AS across all experiments.
GroundingDINO is selected for its strong performance on
dense open-vocabulary detection, while Q-Align is chosen for
its discrete text-defined level design, which produces
stable scalar scores with lower variance than continuous
VLM-based scoring, a critical property for a reliable
termination signal in the iterative critic loop. This
stability advantage is consistent with recent
findings~(\cite{cao2025artimuse}). The critic
detector is intentionally distinct from the evaluation
detector (OWLv2~(\cite{minderer2023scaling})) to prevent the
critic from optimising directly against the evaluation
metric; we fix the GroundingDINO confidence threshold to 0.3
across all experiments for reproducibility. To verify that \model's gains are driven by the iterative
loop architecture rather than a specific component pairing,
we swap each module independently
(\Cref{tab:component_swap}). Replacing GroundingDINO with
OWLv2 in the critic role introduces critic--evaluator
overlap that gives OWLv2 an inherent advantage, yet MAE
remains comparable to the default. Replacing Q-Align with
ImageReward~(\cite{xu2023imagereward}) increases MAE modestly
but still substantially outperforms the best external
baseline (3DIS: 14.55). Even the weakest configuration in
the table beats all published baselines by a wide margin,
confirming that \model~is robust to component choice.
Notably, the $^\dagger$OWLv2-as-critic row is an implicit cross-detector check: despite critic--evaluator overlap giving OWLv2 an inherent advantage, MAE remains comparable to the GroundingDINO default, confirming gains are not a detector artefact.
\begin{figure}[t]
  \centering
  \includegraphics[width=\linewidth]{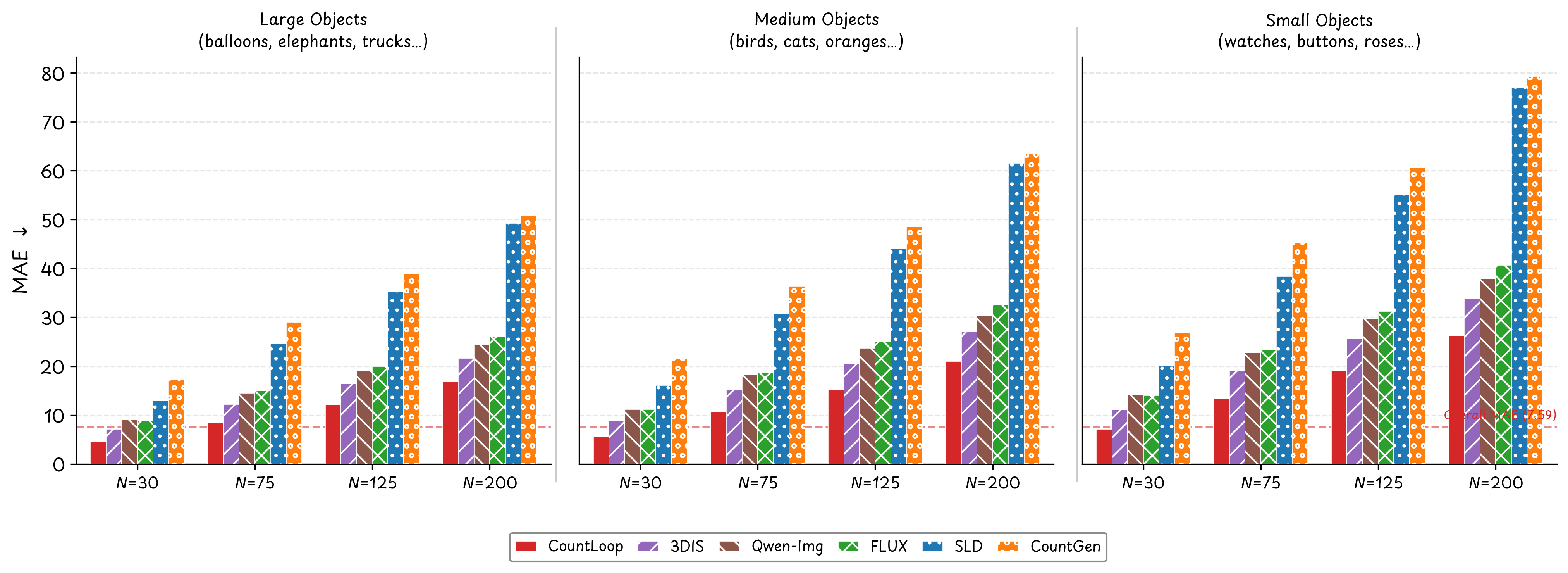}
  \caption{%
    \textbf{Per-regime MAE at representative instance counts on
    \datasetsingle.}
    Categories are grouped by object size (Large / Medium / Small);
    bars show MAE at $N\!\in\!\{30,75,125,200\}$ for all six methods.
    The Large\,$<$\,Medium\,$<$\,Small ordering is method-agnostic,
    reflecting canvas-density pressure and OWLv2 precision loss on small,
    densely-packed instances.
    \model\ (red) achieves the lowest MAE in every regime at every $N$;
    the gap over next-best 3DIS widens with $N$, reaching
    $\Delta$MAE\,$\approx\!7.6$ at $N\!=\!200$ in the small-object regime
    \change{(MAE 26.5 vs.\ 34.1, or 13\% vs.\ 17\% relative MAE)}.
    The dashed reference line marks \model's overall MAE on
    \datasetsingle.}
  \label{fig:regime_bars}
\end{figure}

\myparagraph{Instance Count Scalability by Object Regime}
~\Cref{fig:regime_bars} disaggregates the MAE-$N$ relationship by object-size regime, grouping categories into
\emph{large} (balloons, elephants, trucks), \emph{medium} (birds, cats,
oranges), and \emph{small} (watches, buttons, roses) classes, directly
addressing the question of whether \model\ has a practical upper bound on
reliable instance generation.
\begin{wraptable}{r}{0.45\linewidth}
  \centering
  \caption{\change{Component swap on \datasetsingle. Evaluation
  detector is OWLv2 throughout. Values vs.\ Best Baseline report
  relative MAE reduction over 3DIS (MAE 14.55). $^\dagger$OWLv2 as both
  critic and evaluator gives an inherent advantage, yet
  MAE remains comparable.}}
  \label{tab:component_swap}
  \setlength{\tabcolsep}{3pt}
  \resizebox{\linewidth}{!}{%
    \begin{tabular}{@{}llcc@{}}
      \toprule
      \textbf{Critic OVD} & \textbf{Aesthetic}
        & \textbf{MAE}$\downarrow$
        & \change{\textbf{vs Best}} \\
      \midrule
      \rowcolor{L2IGreen}
GroundingDINO & Q-Align (default) & \textbf{7.59} & +48\% \\
\rowcolor{L2IGreen}
OWLv2$^\dagger$ & Q-Align & 8.57 & +41\% \\
\rowcolor{L2IGreen}
GroundingDINO & ImageReward & 9.21 & +36\% \\
      \bottomrule
    \end{tabular}%
  }
\end{wraptable}
Across all six methods, MAE increases monotonically with $N$, consistent with
the aggregate trend.
The rank ordering Large\,$<$\,Medium\,$<$\,Small is
\emph{method-agnostic}: it reflects two compounding factors independent of
generation strategy.
First, small objects occupy a larger fraction of the canvas at high instance
counts, intensifying cross-attention identity confusion during diffusion
sampling.
Second, OWLv2 localisation precision degrades on densely packed, sub-pixel
instances, introducing systematic upward bias in the detector-based MAE
estimate for all methods equally.
Since both factors are architectural constants of the evaluation setup rather
than properties of any single method, the tier ordering cannot be attributed
to \model's design, and the shared bias does not inflate \model's relative
advantage, as it affects every method in the comparison identically.

Critically, \model~maintains the lowest MAE in every regime at every $N$,
with the advantage \emph{widening} rather than narrowing at high density:
the gap over next-best 3DIS reaches $\Delta$MAE\,$\approx\!7.6$ in the
small-object regime at $N\!=\!200$.
\change{\model~maintains an MAE of 26.5 in the hardest setting (small, $N=200$, a relative MAE of only 13\%)
and 16.5 in the large-object regime (8\% relative MAE), well beyond all prior methods.}
The performance ceiling is \emph{category-dependent} and scales with object
size; \model~raises it substantially across all regimes, with its advantage
widening at the benchmark ceiling of $N\!=\!200$.

\section{Conclusion}

We presented \model, a training-free, iterative framework for high-instance image generation with {significantly improved count fidelity} and strong visual quality. VLM-based planning graphs, instance-driven attention, and cumulative attention composition overcome count saturation, semantic leakage, and rigid layouts, with a critic-in-the-loop refining generation. On COCO-Count, T2I-CompBench, and new high-instance benchmarks, \model~reduces counting error by up to 57\% on standard benchmarks and 43--48\% on high-instance scenes, achieving the highest or comparable spatial quality throughout.
\begin{figure}[!t]
    \vspace{-1em}
    \centering
    \includegraphics[width=0.7\linewidth]{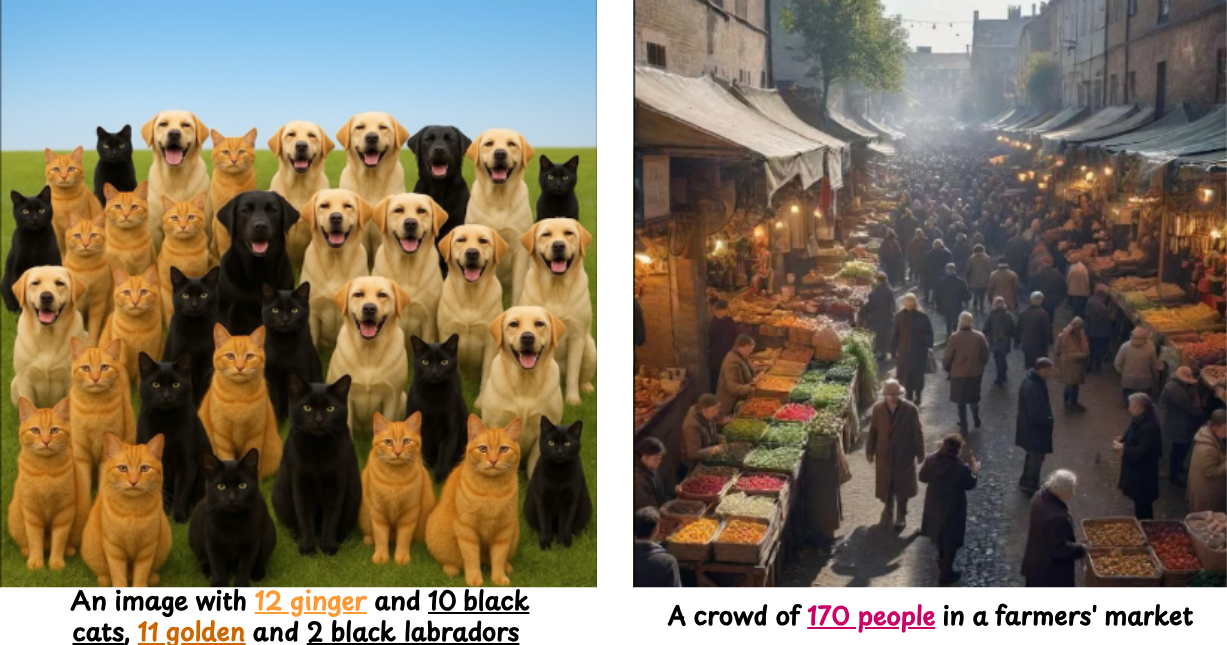}
    \caption{Failure cases}
    \label{fig:limitations}
    \vspace{-1em}
\end{figure}

\myparagraph{Limitations} 
As a training-free system, \model~inherits the limitations of its frozen VLM and detector. Dense occlusions, especially in human scenes, can degrade attention quality and spatial consistency. Lacking explicit 3D priors, \model~struggles with generating objects in different poses and complex perspectives. 
Moreover, strong layout guidance can reduce intra-class diversity by biasing toward canonical poses for \change{count fidelity}. {Furthermore, as our per-tier empirical evidence shows, current capabilities do not establish guaranteed or exact count control at very high densities ($N > 30$), where exact-match rates remain low due to heavy inter-object occlusion, canvas resolution constraints, and the imperfect proxy nature of automated detectors. Achieving guaranteed exact counts in extreme-density regimes remains an open challenge.} Some of these limitations are shown in \Cref{fig:limitations}.

\section{Acknowledgements}
The authors gratefully acknowledge NVIDIA Corporation for support through the NVIDIA Academic Grant Program, which provided computational resources for this research. The authors also acknowledge the use of resources provided by the Isambard-AI National AI Research Resource (AIRR) (\cite{mcintoshsmith2024isambardai}), operated by the University of Bristol and funded by the UK Government’s Department for Science, Innovation and Technology (DSIT) through UK Research and Innovation and the Science and Technology Facilities Council [ST/AIRR/I-A-I/1023].

\bibliographystyle{ieeenat_fullname}
\bibliography{ref}


\end{document}